\documentclass[conference]{IEEEtran}
\IEEEoverridecommandlockouts

\usepackage{cite}
\usepackage{amsmath,amssymb,amsfonts}
\usepackage{algorithmic}
\usepackage{graphicx}
\usepackage{textcomp}
\usepackage{xcolor}
\usepackage{float}
\usepackage{booktabs}
\usepackage{amssymb}
\usepackage{color}
\usepackage{caption}
\usepackage{subfigure}
\usepackage{multirow}
\usepackage{array}
\usepackage{hyperref}
\usepackage{pdfpages}
\usepackage{tabularx}

\def\BibTeX{{\rm B\kern-.05em{\sc i\kern-.025em b}\kern-.08em
    T\kern-.1667em\lower.7ex\hbox{E}\kern-.125emX}}
\begin{document}

\title{Leveraging Pre-Trained Models for Multimodal Class-Incremental Learning under Adaptive Fusion
\thanks{ ${}^{\dag}$ Corresponding author (fmmeng@uestc.edu.cn).} 
}

\author{
    \IEEEauthorblockN{
        Yukun Chen, 
        Zihuan Qiu, 
        Fanman Meng\textsuperscript{\dag}, 
        Hongliang Li,
        Linfeng Xu,
        Qingbo Wu
    }
    
    \IEEEauthorblockA{
        University of Electronic Science and Technology of China, Chengdu, China\\
        \{yukunchen@std.,zihuanqiu@std.,fmmeng@,hlli@,lfxu@,qbwu@\}uestc.edu.cn
    }
}

\maketitle

\begin{abstract}
Unlike traditional Multimodal Class-Incremental Learning (MCIL) methods that focus only on vision and text, this paper explores MCIL across vision, audio and text modalities, addressing challenges in integrating complementary information and mitigating catastrophic forgetting. To tackle these issues, we propose an MCIL method based on multimodal pre-trained models. Firstly, a Multimodal Incremental Feature Extractor (MIFE) based on Mixture-of-Experts (MoE) structure is introduced  to achieve effective incremental fine-tuning for AudioCLIP. Secondly, to enhance feature discriminability and generalization, we propose an Adaptive Audio-Visual Fusion Module (AAVFM) that includes a masking threshold mechanism and a dynamic feature fusion mechanism, along with a strategy to enhance text diversity. Thirdly, a novel multimodal class-incremental contrastive training loss is proposed to optimize cross-modal alignment in MCIL. Finally, two MCIL-specific evaluation metrics are introduced for comprehensive assessment. Extensive experiments on three multimodal datasets validate the effectiveness of our method.

\end{abstract}


\section{Introduction}
\label{Introduction}
In real-world applications, data often changes dynamically, requiring models trained with deep learning techniques to adapt to continuously arriving data of new classes. Class-Incremental Learning (CIL) aims to continually learn new class knowledge without catastrophic forgetting of previously learned classes.

Three types of methods such as replay-based methods \cite{e1,e2,e3,e4,e5}, regularization-based methods \cite{e6,e7,e8,e9,e10}, and architecture-based methods \cite{e11,e12,e13,e14,e15} have been proposed to keep the recognition of old class. Recently, pre-trained models with parameter-efficient fine-tuning techniques \cite{e24,e25,e26,e27,e28} is widely used for Class-Incremental Learning (CIL) \cite{e16,e17,e18,e19,e20,e21,e22}, achieving promising results in addressing catastrophic forgetting due to the better generalization of the pre-trained model.

\begin{figure}[t]
    \centering
    \subfigure[]{\includegraphics[width=0.95\linewidth]{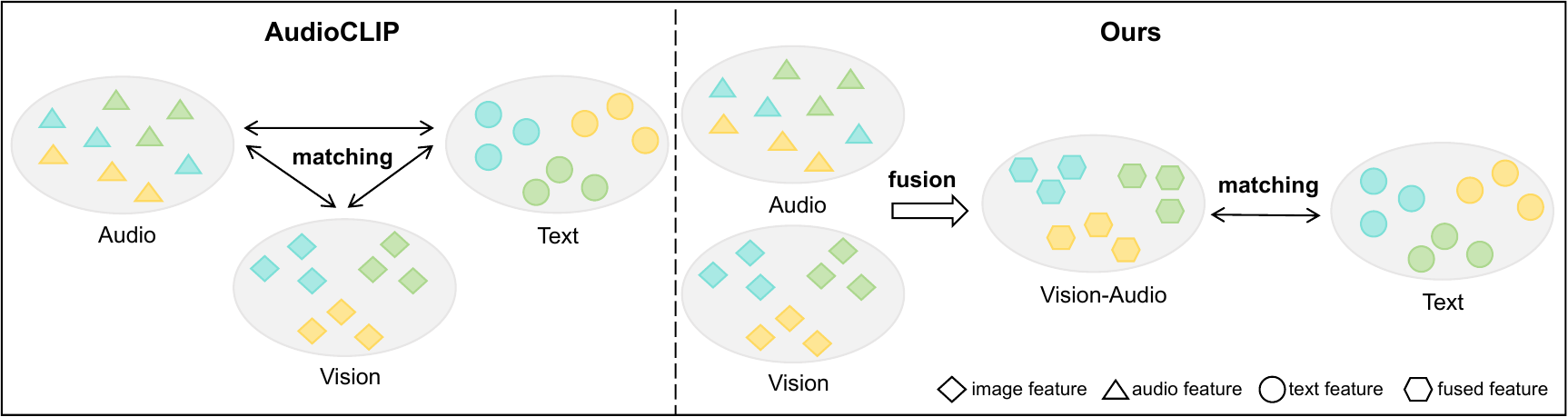}\label{fig:f1a}}
    \subfigure[]{\includegraphics[width=0.4\linewidth]{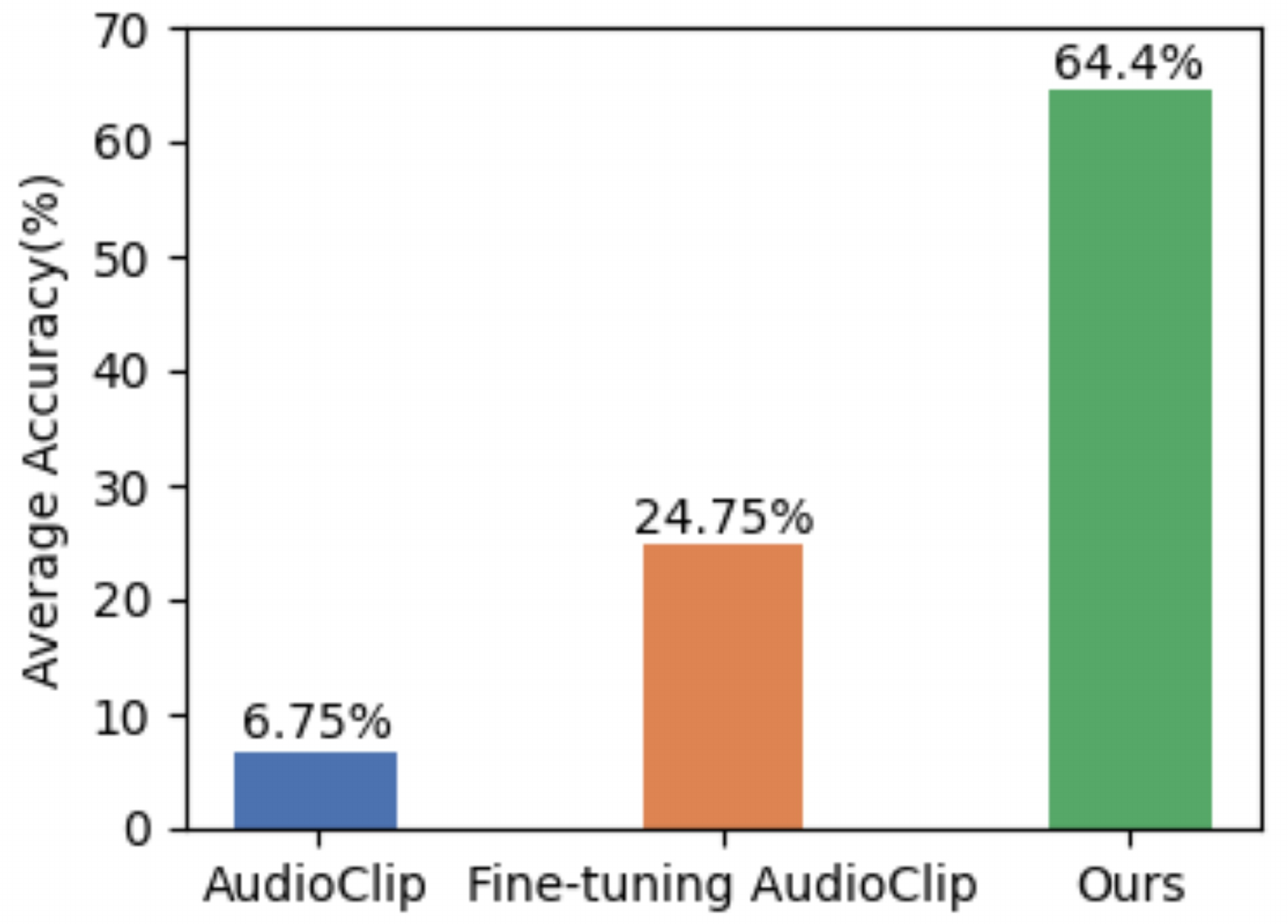}\label{fig:f1b}}
    \subfigure[]{\includegraphics[width=0.4\linewidth]{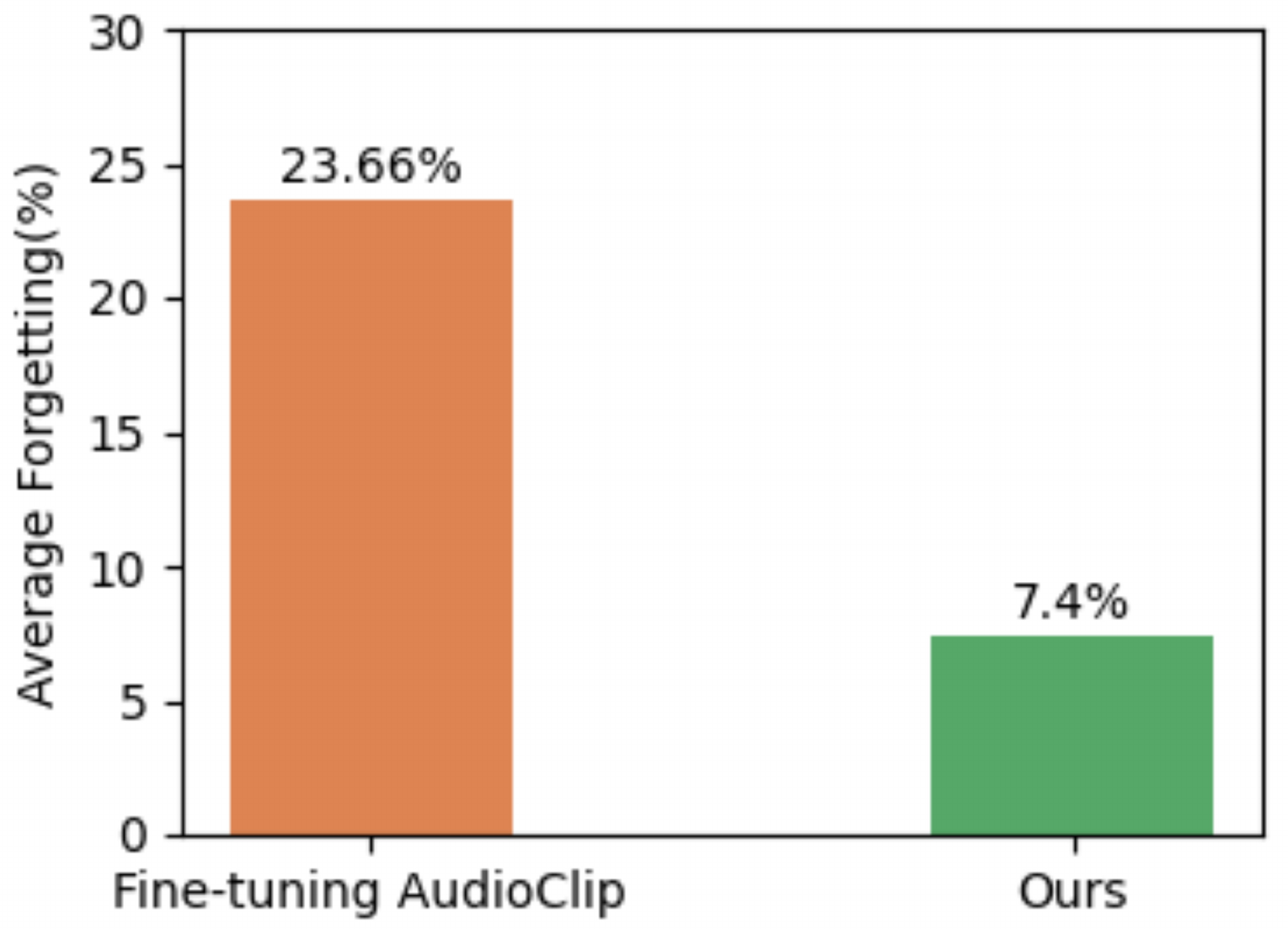}\label{fig:f1c}}
    \caption{(a) Illustration of cross-modal matching in the feature space for AudioCLIP and our method. (b) and (c) show the comparison of classification performance across different methods under the MCIL setting, using average Top-1 accuracy and average forgetting as evaluation metrics, respectively.}
    \label{fig:f1}
\end{figure}

\begin{figure*}[t]  
    \centering  
    \includegraphics[scale=0.48]{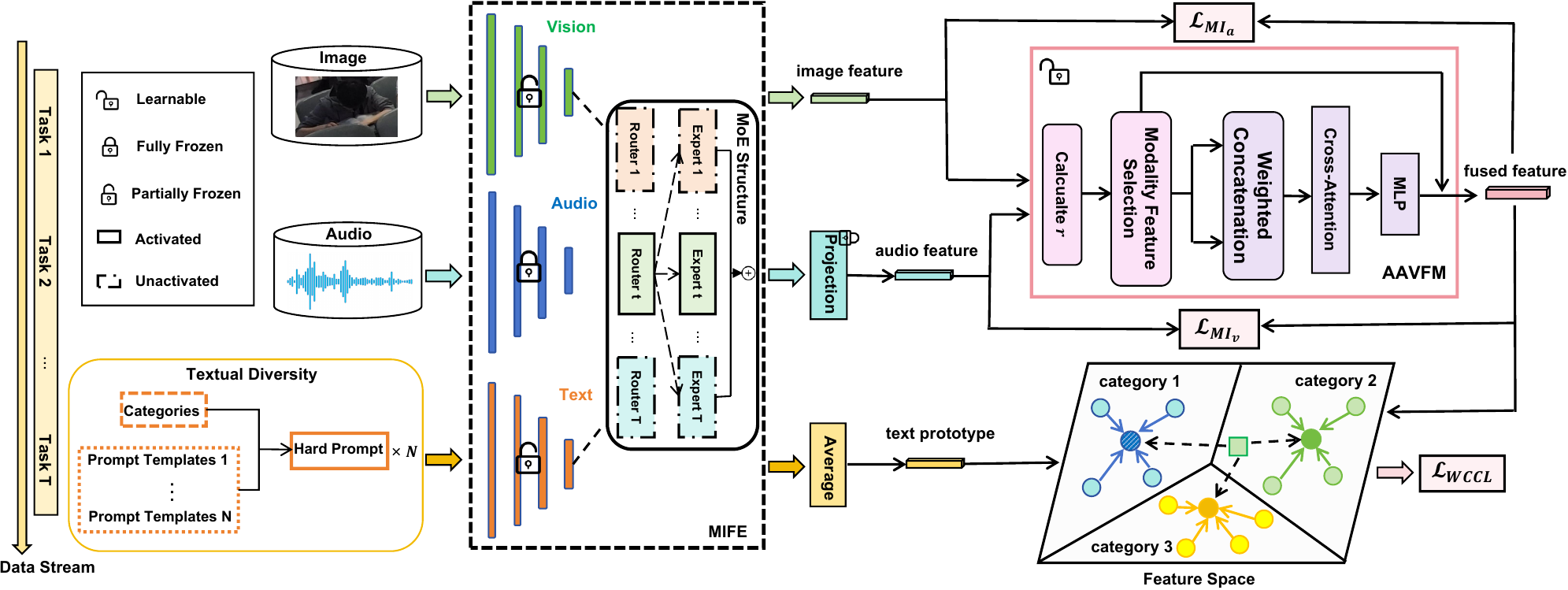}  
    \captionsetup{font=small, labelfont=bf}  
    \caption{Overall framework of our proposed method.}
    \label{fig:f2}  
\end{figure*}

However, these methods are limited to visual and text modalities. In practical applications such as classroom settings, data often appear in multiple modalities, such as visual, audio, and text. This leads to a more difficult task of Multimodal Class-Incremental Learning (MCIL) \cite{e29,e30} with more than three modalities, which receives little attention. The challenge here is how to handle the vision, audio and text modalities, which differ significantly from each other.

Multimodal pre-trained model AudioCLIP \cite{e31} possesses valuable prior knowledge of multiple modalities, offering a potential solution for MCIL. Meanwhile, as shown in Fig. \ref{fig:f1}, the AudioCLIP-based method has two main limitations in MCIL: poor generalization and significant performance degradation during incremental fine-tuning. The reasons are as follows. Based on a contrastive learning framework, AudioCLIP considers each modality equally during the cross-modal alignment and neglects the differences of the quality in modalities present in practical data, which limits the model’s generalization ability \cite{e32}. Moreover, AudioCLIP lacks an effective incremental fine-tuning strategy to balance the retention of knowledge from old classes with the learning of new ones, which leads to the forgetting of old classes.

In this paper, we propose an MCIL method based on multimodal pre-trained models, specifically addressing visual, audio and text modalities. To tackle the first limitation above, we propose a multimodal fusion approach that efficiently combines the modalities of different quality by designing an Adaptive Audio-Visual Fusion Module (AAVFM) with a masking threshold mechanism and a dynamic feature fusion mechanism. This module first adaptively integrates audio and visual modality data, then the obtained feature with clear classification boundaries are matched with text features obtained by a strategy for enhancing text diversity. To address the second limitation, a Multimodal Incremental Feature Extractor (MIFE) based on Mixture-of-Experts (MoE) structure \cite{e22,e33} is proposed to achieve effective incremental fine-tuning for AudioCLIP. Additionally, a loss function based on cross entropy loss is proposed for multimodal class-incremental contrastive training, along with two novel evaluation metrics for MCIL. Extensive experiments on miniARIC, ImageNet-ESC-19 and ImageNet-ESC-27 datasets demonstrate the effectiveness of our method.

\section{PROPOSED METHOD}
\label{sec:format}
In this section, we present the proposed method, with the overall framework illustrated in Fig. \ref{fig:f2}.

Regarding the problem formulation in MCIL, given a set of tasks \(\{\mathcal{T}_t\}_{t=1}^T\), the model needs to sequentially access and learn \(T\) tasks involving multimodal data such as visual, audio, and text. During the training phase, the data accessible to the model while learning task \(t\) is \(\mathcal{D}_t = \left\{\left(v_i, a_i, l_i\right)\right\}_{i=1}^{n_t}\), where \(v_i\) denotes the input image, \(a_i\) denotes the input audio matching \(v_i\), \(l_i\) is the corresponding label, and \(n_{t}\) is the number of samples in task \(t\). We use \(\mathcal{C}_{t}\) to denote the new classes in task \(t\), and \(\mid \mathcal{C}_{1:t}\mid=\sum_{t^{\prime}=1}^t \mid\mathcal{C}_{t^{\prime}}\mid\) to denote all classes the model has seen up to task \(t\). After completing training on \(\mathcal{C}_{t}\), the model will be evaluated on all previously learned tasks. Since the task identity \(t\) is unknown during inference, the model faces the challenge of distinguishing all previously encountered classes \(\mathcal{C}_{1:t}\).


\subsection{Multimodal Incremental Feature Extractor}
\label{ssec:subhead2}
To address the issue of diversity mismatch between the text and the visual and audio modalities, we use a set of text prompt templates \(\{PT|_{l=1}^N\}\) to expand each category label into \(N\) hard prompts containing more semantic knowledge. This approach increases text diversity, providing the model with richer contextual information and making it easier for the model to understand and distinguish between different classes.

A Multimodal Incremental Feature Extractor (MIFE) is constructed with three parallel feature extractors \((\mathcal{F}_v,\mathcal{F}_a,\mathcal{F}_l)\). MIFE is used to extract features \(\left[\mathcal{F}_{v}(v_{t}),\mathcal{F}_{a}(a_{t}),\overline{{\mathcal{F}_{l}(l_{t})}}\right]\) from the multimodal input \(x_t=(v_t,a_t,l_t)\). The audio feature extractor \(\mathcal{F}_a\) uses the frozen audio encoder from the pre-trained AudioCLIP model. The visual feature extractor \(\mathcal{F}_v\) and the text feature extractor \(\mathcal{F}_l\) 
incorporate a MoE structure. This MoE structure is added to each Transformer block in the pre-trained AudioCLIP’s image and text encoders, respectively. During training, all components of \(\mathcal{F}_v\) and \(\mathcal{F}_l\) remain frozen except for the MoE structure, enabling parameter-efficient incremental fine-tuning.

The MoE structure consists of task-specific experts and routers. Each expert learns knowledge for incremental tasks, and the router controls their state, determining whether they are active or inactive. Active experts are updated, while inactive ones are frozen. The router uses gating weights to integrate the outputs of the experts into the final MoE output. This strategy effectively prevents forgetting old knowledge while enabling knowledge sharing. We use a parameter-efficient adapter as the expert, with its parameters decomposed using LoRA \cite{e34}. The router is a single MLP \cite{e35}.

\subsection{Adaptive Audio-Visual Fusion Module}
\label{ssec:subhead3}
In multimodal datasets, the data quality across different modalities often varies. We classify modalities with higher data quality as strong modalities and those with lower quality as weak modalities. To overcome the limitations of unimodal and fully leverage the complementary information from different modalities, we designed the Adaptive Audio-Visual Fusion Model (AAVFM). Firstly, in AAVFM, we set a masking threshold \(th\) to filter out noise interference from the weak modality. The pearson correlation coefficient \(r\) between visual feature \(\mathcal{F}_{v}\left(v_{t}\right)\) and audio feature \(\mathcal{F}_{a}\left(a_{t}\right)\) is calculated using the following formula:
\begin{equation}
r=\frac{\sum_{i=1}^n\Delta v_i\Delta a_i}{\sqrt{\sum_{i=1}^n\Delta v_i^2}\sqrt{\sum_{i=1}^n\Delta a_i^2}}
\label{eq:eq1}
\end{equation}
where \(\Delta v_i=\mathcal{F}_v(v_t)_i-\mu_v\) and \(\Delta a_i=\mathcal{F}_a(a_t)_i-\mu_a\), with \(\mu_v\) and \(\mu_a\) representing the means of \(\mathcal{F}_v(v_t)_i\) and \(\mathcal{F}_a(a_t)_i\) respectively. If \(r<th\), only the features from the strong modality are used. If \(r\geq th\), features from both modalities are utilized and fused. Additionally, the feature fusion module in AAVFM incorporates learnable modality weight parameters and a cross-attention mechanism. A frozen linear layer \(\mathcal{P} \in \mathbb{R}^{1024 \times 512}\) is used to project \(\mathcal{F}_{a}\left(a_{t}\right)\) into the embedding space of \(\mathcal{F}_{v}\left(v_{t}\right)\). Learnable weight parameters are set for both the visual and audio modalities to perform weighted concatenation of \(\mathcal{F}_{v}\left(v_{t}\right)\) and \(\mathcal{P}(\mathcal{F}_a(a_t))\) to obtain \(f_t^{\mathrm{concat}}\). Cross-attention is then applied to \(\mathcal{F}_{v}\left(v_{t}\right)\) and \(f_t^{\mathrm{concat}}\):
\begin{equation}
f_t^{\text{c-a}} = Softmax\left(\frac{\mathcal{F}_{v}\left(v_{t}\right) {f_t^{\text{concat}}}^\top}{\sqrt{d_k}}\right) f_t^{\text{concat}}
\end{equation}
where \(Softmax(\cdot)\) is used to convert scores into a probability distribution, while \(d_k\) serves as a scaling factor.

After inputting \(f_t^{\mathrm{c-a}}\) into a trainable MLP, the final fused feature \(f_t^{\mathrm{fusion}}\) is obtained. This feature fusion module allows the model adaptively adjust the contributions of different modalities, resulting in \(f_t^{\mathrm{fusion}}\) that contains more effective information.

\subsection{Loss Function}
\label{ssec:subhead4}
We obtain the probability prediction distribution based on the cosine similarity between fused feature \(f_t^{\mathrm{fusion}}\) and text prototype \(\overline{{\mathcal{F}_{l}(l_{t})}}\):

\begin{equation}\tilde{y}_t=Softmax\Big(sim\Big(f_t^{\mathrm{fusion}},\overline{\mathcal{F}_l(l_t)}\Big)\Big)\end{equation}
where \(sim(\cdot,\cdot)\) calculates the cosine similarity between two feature vectors.

To improve cross-modal alignment, we extend the standard cross entropy loss \(\mathcal{L}_{\mathrm{CE}}\) by introducing a weighting factor \(w_{ij}\), which measures the similarity between samples within the visual modality. \(\mathcal{L}_{\mathrm{CW}}\) enables a more fine-grained calculation of modality gaps and is defined as follows:
\begin{equation}
w_{ij}=\frac12sim\left(\mathcal{F}_v(v_t)_i,\mathcal{F}_v(v_t)_j\right)+\frac12
\end{equation}
\begin{equation}
\mathcal{L}_{\mathrm{CW}}=\frac1n\sum_{i=1}^n\frac1n\sum_{j=1}^nw_{ij}\cdot\mathcal{L}_{\mathrm{CE}}
\end{equation}

Additionally, we introduce \(\mathcal{L}_{\mathrm{MI}}\) to maximize the mutual information between \(f_t^{\mathrm{fusion}}\) and both \(\mathcal{F}_{v}(v_t)\) and \(\mathcal{F}_{a}(a_t)\). This helps \(f_t^{\mathrm{fusion}}\) capture detailed information from different modalities and minimize information loss. The definition is as follows:
\begin{equation}
\mathcal{L}_{\mathrm{MI_v}} = -I(f_t^{\mathrm{fusion}}; \mathcal{F}_v(v_t))
\end{equation}
\begin{equation}
\mathcal{L}_{\mathrm{MI_a}} = -I(f_t^{\mathrm{fusion}}; \mathcal{F}_a(a_t))
\end{equation}
\begin{equation}\mathcal{L}_{\mathrm{MI}}=\mathcal{L}_{\mathrm{MI_v}}+\mathcal{L}_{\mathrm{MI_a}}\end{equation}
where \(I(\cdot;\cdot)\) denotes mutual information, which is used to measure the degree of information shared between two feature vectors.
In summary, the total loss function is defined as:
\begin{equation}\mathcal{L}=\alpha\mathcal{L}_{\mathrm{CW}}+(1-\alpha)\mathcal{L}_{\mathrm{MI}}\label{eq:eq6}\end{equation}
where \(\alpha\in[0,1]\) is a hyperparameter.

\subsection{Evaluation Metric}
\label{ssec:subhead4}
To evaluate our method on MCIL and address the limitations of existing metrics that only assess model performance in a single modality or aspect, we propose two novel evaluation metrics, \(M_1\) and \(M_2\), motivated by the methods in \cite{e22,e36}.

In metric \(M_1\), a task similarity measure \(w_t\), which accounts for both semantic and feature distribution levels, is introduced. \(w_t\) is used to weight the forgetting \(For_{t}\) at each incremental stage, and combined with the average Top-1 accuracy \(Acc_{\mathrm{avg}}\) over the entire MCIL process to jointly assess the model's robustness and overall performance in handling different incremental tasks. The formula for calculating \(M_1\) is as follows:
\begin{equation}
M_1=\frac{1}{2}Acc_{\mathrm{avg}}+\frac{1}{2T}\sum_{t=1}^{T}\left(1-w_t\right)\left(100-For_{t}\right)
\end{equation}
where, \(w_t\) is obtained by linearly mapping the cosine similarity and \(w_t\in[0,1]\). The range of this metric is \(M_1\in[0,100]\), expressed as a percentage.

Metric \(M_2\) evaluates the model's capability in MCIL from three aspects: memory stability \(BWT_t\), learning plasticity \(FWT_t\) and overall classification performance \(Acc_{\mathrm{avg}}\). It also assesses the model's utilization of multimodal information by measuring the richness of fused features and the consistency of multimodal information using normalized mutual information \(NMI_{\mathrm{f-v}}\) and \(NMI_{\mathrm{f-a}}\). The range of this metric is \(M_2\in[0,1]\). The formula for calculating \(M_2\) is as follows:
\begin{equation}
\begin{split}
M_2=\frac14Acc_{\mathrm{avg}}+\frac1{4T}\sum_{t=1}^T\left(BWT_t+FWT_t\right)\\
+\frac14(NMI_{\mathrm{f-v}}+NMI_{\mathrm{f-a}})
\end{split}
\end{equation}

\begin{table}[t]
\captionsetup{font=small, labelfont=bf}  
\caption{
Performance of various methods across different task settings on the three multimodal datasets, measured by average Top-1 accuracy throughout the entire MCIL process. \textbf{Best results - in bold.}}
\centering
\scalebox{0.9}{
\begin{tabular}{l|cccccc}  
\hline
\multirow{2}*{Method} & \multicolumn{2}{c}{miniARIC} & \multicolumn{2}{c}{ImageNet-ESC-19} & \multicolumn{2}{c}{ImageNet-ESC-27} \\ 
\cline{2-7}
& \(T=3\) & \(T=5\) & \(T=3\) & \(T=6\) & \(T=5\) & \(T=8\) \\
\hline
Zero-shot\cite{e31} & 6.94 & 6.75 & 32.49 & 35.44 & 22.22 & 17.36 \\
Partial Fine-tune & 27.02 & 24.75 & 69.88 & 74.52 & 64.70 & 68.01 \\
LwF\cite{e6} & 27.95 & 25.17 & 71.43 & 79.08 & 71.32 & 74.88 \\
iCaRL\cite{e1} & 31.93 & 26.76 & 81.76 & 77.96 & 80.30 & 74.09 \\
TOPIC\cite{e12} & 28.37 & 27.95 & 76.81 & 79.27 & 73.95 & 72.95 \\
MoE-CLIP\cite{e22} & 49.19 & 40.01 & 92.60 & 94.52 & 95.05 & 93.39 \\
Ours & \textbf{67.76} & \textbf{64.40} & \textbf{95.37} & \textbf{95.96} & \textbf{96.74} & \textbf{96.02} \\
\hline
\end{tabular}}
\label{tab:table1}
\end{table}

\section{Experiments}
\label{sec:Experiments}
\subsection{Experimental Setting}
\label{ssec:subhead5}
We conduct experiments on three datasets: miniARIC \cite{e37}, ImageNet-ESC-19 and ImageNet-ESC-27 \cite{e38} under the MCIL setting. The miniARIC dataset consists of behavior classes of teachers and students across 33 classroom scenes. It not only contains many challenging samples for classification but also has lower data quality in the audio modality. We randomly sample images from the miniARIC dataset and crop them using bounding boxes as visual modality samples. For the audio data, we extract Mel-frequency cepstral coefficient (MFCC) features. The text modality is generated by expanding class names with fixed templates. The ImageNet-ESC-19 and ImageNet-ESC-27 datasets follow the setup in \cite{e38}. All experiments are conducted on an NVIDIA GeForce RTX 4090 GPU using pre-trained AudioCLIP, based on ViT-B/16 \cite{e39} and ESResNeXt \cite{e40}, as the backbone. Training is performed with the AdamW optimizer \cite{e41} and CosineAnnealingLR scheduler \cite{e42}. We set the number of text prompt templates to \(N=35\), the threshold \(th\) in AAVFM to 0.8 and \(\alpha\) in \(\mathcal{L}\) to 0.7. 

\begin{figure}[t]
    \centering
    \subfigure[]{
        \centering
        \includegraphics[scale=0.14]{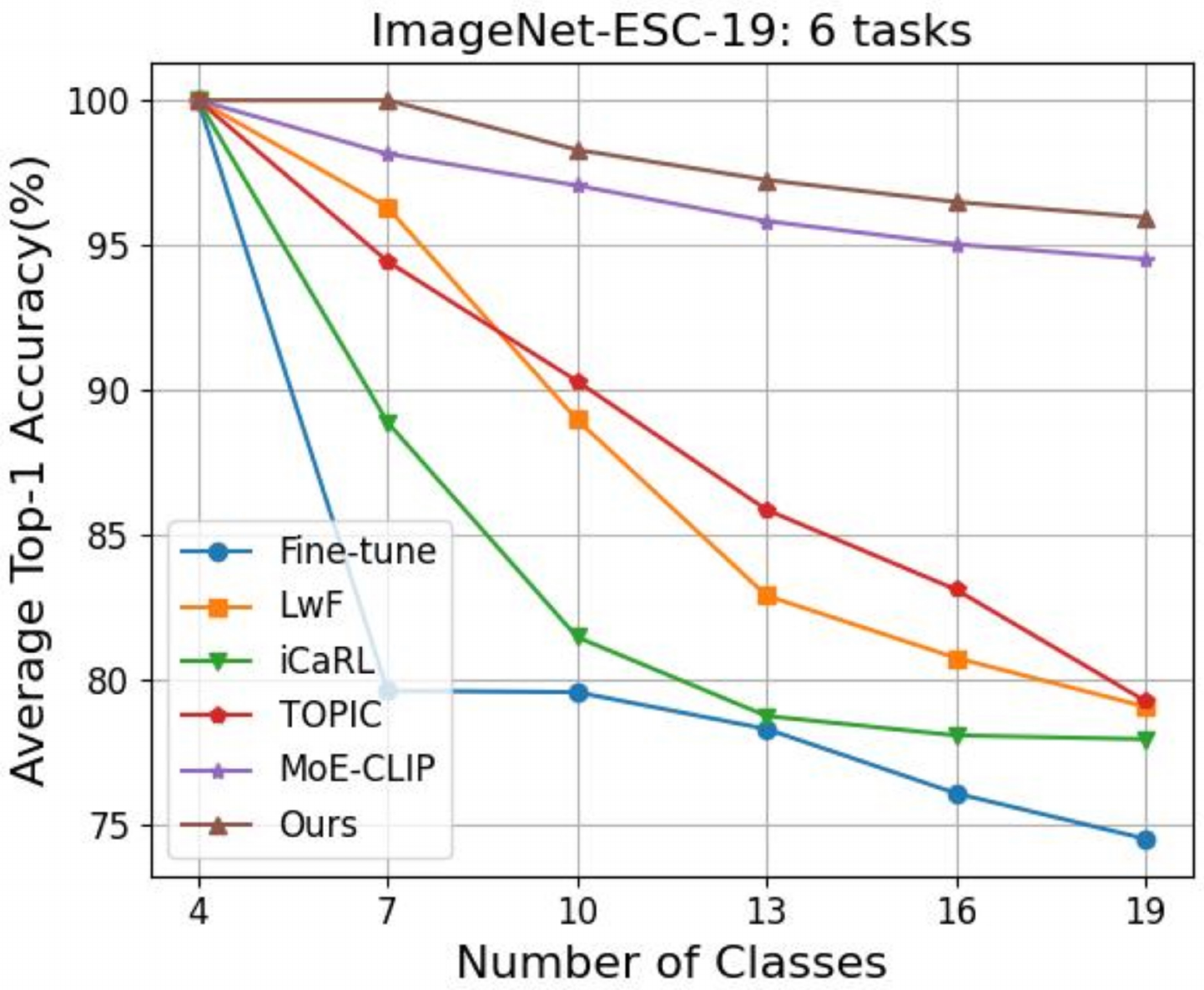}
        \label{fig:fig4}
    }
    \subfigure[]{
        \centering
        \includegraphics[scale=0.16]{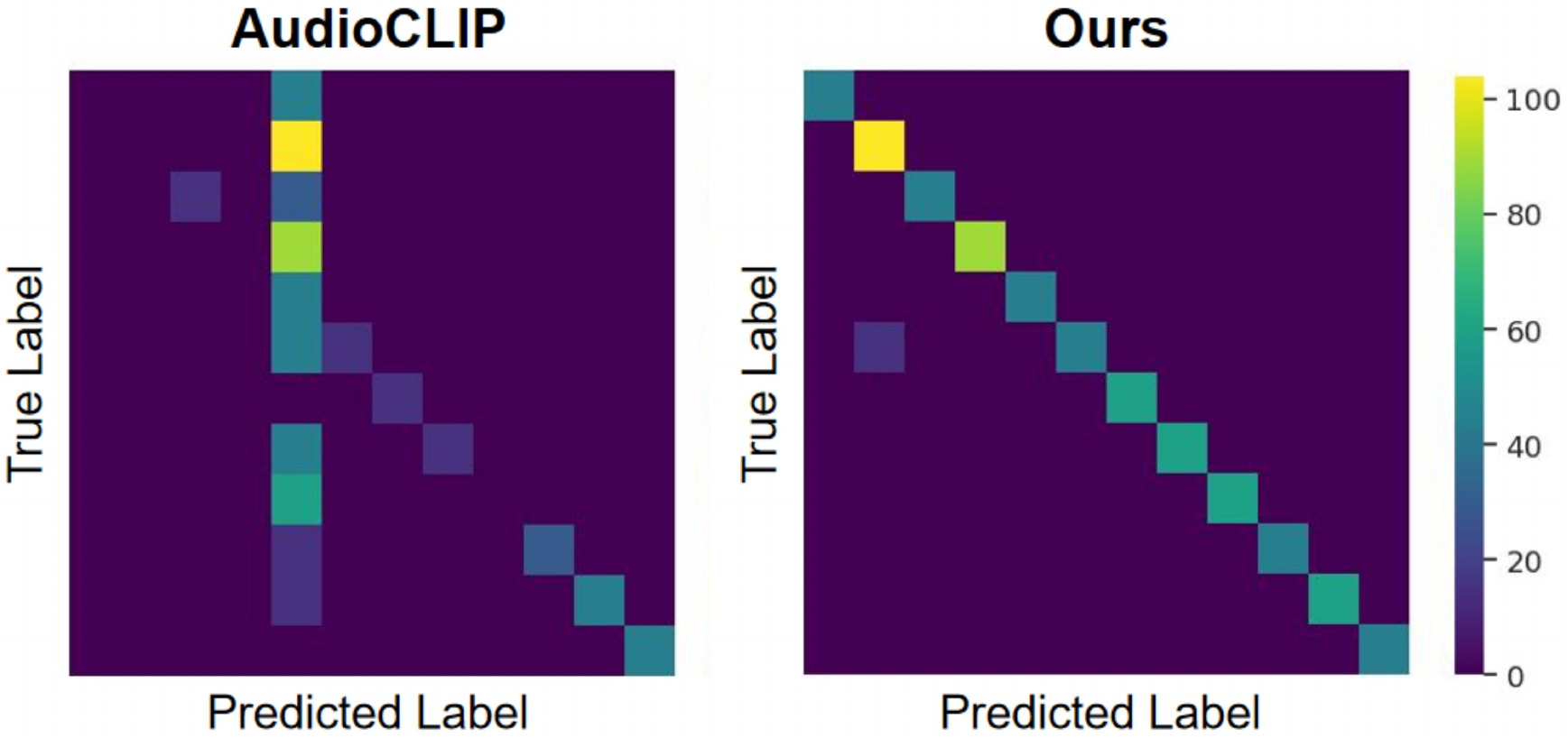}
        \label{fig:fig8}
    }
    \captionsetup{font=small, labelfont=bf}  
    \caption{(a) Forgetting curves formed by the average Top-1 accuracy at each incremental step. (b) Comparison of confusion matrices generated during the first incremental stage.}
    \label{fig:fig6}
\end{figure}

\begin{table}[t]
    \captionsetup{font=small, labelfont=bf}  
    \caption{Ablation study of different components for our method. \textbf{Best results - in bold.}} 
    \centering
    \scalebox{0.9}{
    \begin{tabular}{l|cccc}
        \toprule
        Method & Avg. Acc. & Last. Acc. & \(M_1\) & \(M_2\) \\
        \hline 
        Baseline & 38.06 & 19.41 & 27.84 & 0.61 \\
        \hspace{0.3cm} + Textual Diversity & 58.73 & 46.86 & 30.51 & 0.74 \\
        \hspace{0.3cm} + AAVFM & 42.44 & 25.91 & 32.35 & 0.64 \\
        \hspace{0.3cm} + $\mathcal{L}$ & 42.86 & 25.84 & 32.30 & 0.64 \\
        Ours & \textbf{64.40} & \textbf{57.59} & \textbf{35.13} & \textbf{0.77} \\
        \bottomrule
    \end{tabular}}
    \label{tab:table2}
\end{table}

\begin{figure}[t]  
    \centering  
    \includegraphics[scale=0.2]{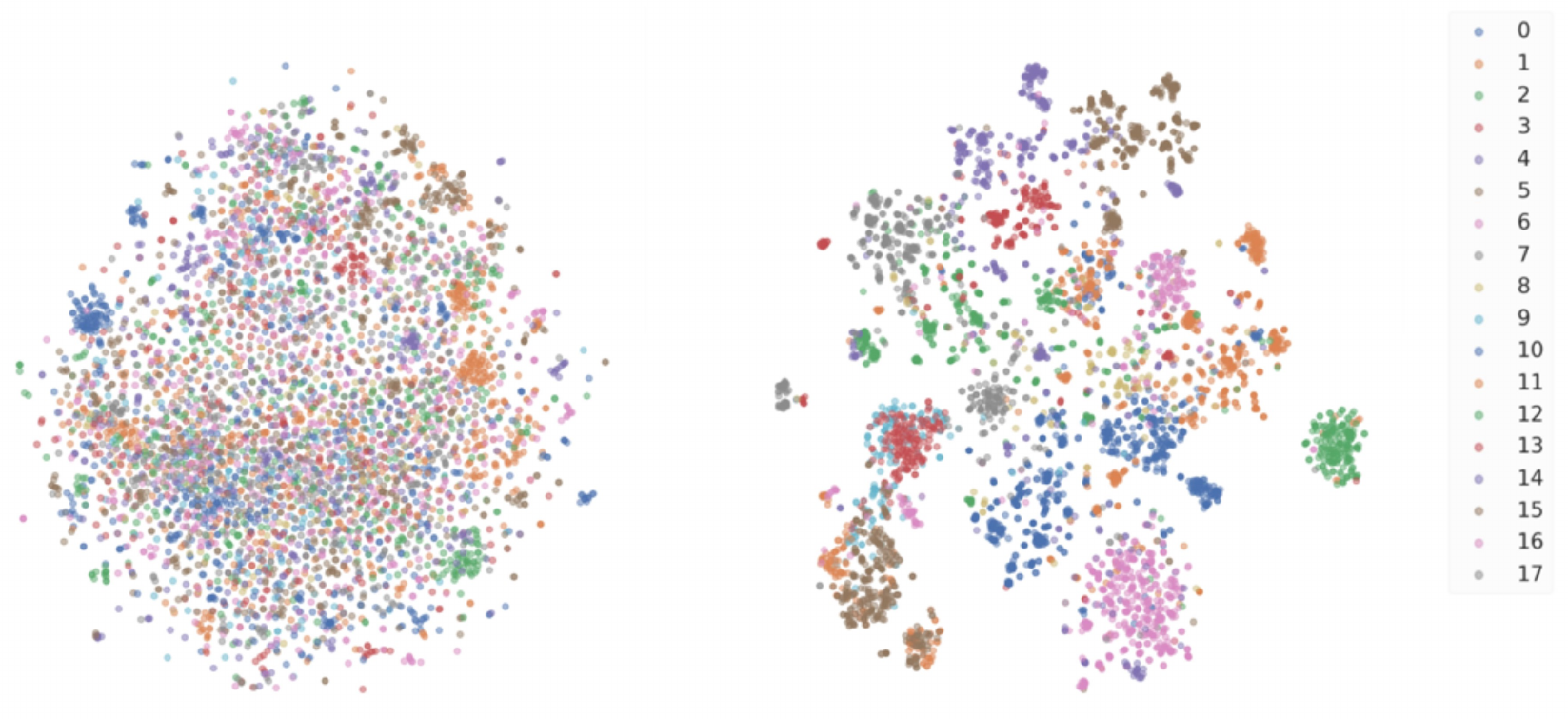}
    \captionsetup{font=small, labelfont=bf}  
    \caption{t-SNE visualization in the first incremental stage.}
    \label{fig:fig7}  
\end{figure}

\begin{table}[t]
\captionsetup{font=small, labelfont=bf}  
\caption{
Ablation study of the AAVFM. \textbf{Best results - in bold.}}
\centering
\scalebox{0.9}{
\begin{tabular}{l|cccc}  
\hline
\multirow{2}*{Method} & \multicolumn{2}{c}{ImageNet-ESC-19} & \multicolumn{2}{c}{ImageNet-ESC-27} \\ 
\cline{2-5}
& Avg. Acc. & Last. Acc. & Avg. Acc. & Last. Acc. \\
\hline
Partial Fine-tune & 74.52 & 66.67 & 64.70 & 61.74 \\
Baseline \textit{w/ AAVFM } & \textbf{95.12} & \textbf{92.00} & \textbf{95.70} & \textbf{91.30} \\
\hline
\end{tabular}}
\label{tab:table3}
\end{table}

\subsection{Comparison with Benchmarks}
\label{ssec:subhead6}
To validate the effectiveness of our method, we conducted multiple evaluations under the MCIL setting. We used task settings of \(T=\{3,5\}\), \(T=\{3,6\}\) and \(T=\{5,8\}\) on the miniARIC, ImageNet-ESC-19, and ImageNet-ESC-27 datasets, respectively. Several classic and state-of-the-art methods in class-incremental learning were re-implemented on AudioCLIP for comparison. Table \ref{tab:table1} presents the results of the comparative experiments, showing that our method significantly outperforms others in MCIL, particularly for challenging samples and when dealing with modality quality imbalance. Fig. \ref{fig:fig4} visualizes the forgetting curves, highlighting our method's superior resistance to forgetting and consistent classification performance. Fig. \ref{fig:fig8} provides confusion matrices on the miniARIC dataset under the \(T=5\) task setting, demonstrating that our method outperforms AudioCLIP in classification accuracy for both new and old classes.

\subsection{Ablation and Analysis Study}
\label{ssec:subhead7}
We conducted an ablation study on the miniARIC dataset under the \(T=5\) task setting to evaluate our method. We removed the textual diversity, replaced the loss function \(\mathcal{L}\) with \(\mathcal{L}_{\mathrm{CE}}\) and used a simple concatenation operation instead of AAVFM as the baseline. 
The experimental results, shown in Table \ref{tab:table2}, demonstrate that each component of our method effectively improves classification performance in MCIL across average Top-1 accuracy, last Top-1 accuracy and two newly proposed MCIL evaluation metrics. The best performance is achieved when all components are combined. 

We performed t-SNE visualization on the visual modality samples during the incremental process. In Fig. \ref{fig:fig7}, the left image shows the results using the default encoder of AudioCLIP, while the right image uses MIFE. The visualization indicates that the features extracted by MIFE exhibit better clustering in the feature space, with greater separability between different classes and clearer classification boundaries.
To validate the effectiveness of our multimodal fusion strategy, we conducted experiments on the ImageNet-ESC-19 and ImageNet-ESC-27 datasets under task settings \(T=6\) and \(T=5\), respectively. As shown in Table \ref{tab:table3}, the experimental results demonstrate that our multimodal fusion strategy using AAVFM significantly improves classification performance in MCIL compared to AudioCLIP's pairwise feature matching strategy.

\section{Conclusion}
\label{sec:Conclusion}
This paper presents a novel MCIL method based on multimodal pre-trained models, integrating the ability to achieve stability-plasticity balance with the capability to merge information from multi-source heterogeneous data into a unified model. Additionally, two new evaluation metrics for MCIL are proposed. To the best of our knowledge, this is the first work to apply multimodal pre-trained models to continual classification tasks. Our method achieves state-of-the-art results across various task settings on three multimodal datasets, providing new perspectives for MCIL research. 
Future work will focus on extending this method to address MCIL challenges in scenarios involving missing modalities.

\section*{Acknowledgment}
This work was supported in part by National Science and Technology Major Project (2021ZD0112001), National Natural Science Foundation of China (Grant 62271119), Natural Science Foundation of Sichuan Province (2023NSFSC1972), and the Independent Research Project of the Civil Aviation Flight Technology and Flight Safety Key Laboratory (FZ2022ZZ06).

\newpage
\bibliographystyle{IEEEtran}
\bibliography{strings1}

\end{document}